\documentclass[accepted]{uai2024}

\usepackage[dvipsnames]{xcolor}
\usepackage[american]{babel}
\usepackage{natbib}
\usepackage{mathtools}
\usepackage{booktabs}
\usepackage{graphicx}
\usepackage{adjustbox}
\usepackage{amsmath,amsfonts,bm}
\usepackage{times}
\usepackage{latexsym}
\usepackage{microtype}
\usepackage{inconsolata}
\usepackage{multirow}
\usepackage{amssymb}
\usepackage{amsthm}
\usepackage{xr}
\usepackage{bbm}
\usepackage{algorithm}
\usepackage{siunitx}
\usepackage{pifont}
\usepackage{subcaption}
\usepackage[noend]{algorithmic}
\usepackage{hyperref}
\usepackage{url}
\usepackage{lineno} 

\newcommand{\cmark}{\textcolor{green}{\ding{51}}}
\newcommand{\xmark}{\textcolor{red}{\ding{55}}}

\title{Knowledge Editing in Language Models via Adapted Direct Preference Optimization}

\author[1*]{Amit Rozner}
\author[1*]{Barak Battash}
\author[2]{Lior Wolf}
\author[1]{\href{mailto:ofirlin@gmail.com?Subject=Knowledge Editing in Language Models via Adapted Direct Preference Optimization}{Ofir Lindenbaum}}
\affil[1]{%
    Faculty of Engineering, Bar Ilan University, Ramat-Gan, Israel
}
\affil[2]{%
    School of Computer Science, Tel Aviv University, Tel-Aviv, Israel
}
\affil[*]{%
These authors contributed equally
}

\begin{document}
\maketitle
\begin{abstract}
Large Language Models (LLMs) can become outdated over time as they may lack updated world knowledge, leading to factual knowledge errors and gaps. Knowledge Editing (KE) aims to overcome this challenge using weight updates that do not require expensive retraining. We propose treating KE as an LLM alignment problem. Toward this goal, we introduce Knowledge Direct Preference Optimization (KDPO), a variation of the Direct Preference Optimization (DPO) that is more effective for knowledge modifications. Our method is based on an online approach that continually updates the knowledge stored in the model. We use the current knowledge as a negative sample and the new knowledge we want to introduce as a positive sample in a process called DPO. We also use teacher-forcing for negative sample generation and optimize using the positive sample, which helps maintain localized changes. We tested our KE method on various datasets and models, comparing it to several cutting-edge methods, with 100 and 500 sequential edits. Additionally, we conducted an ablation study comparing our method to the standard DPO approach. Our experimental results show that our modified DPO method allows for more refined KE, achieving similar or better performance compared to previous methods. 
\end{abstract}

\section{Introduction}
Large language models (LLMs) have achieved remarkable success in various machine learning tasks and are commonly used as foundational models in multiple applications. Training such models \citep{touvron2023llama,touvron2023llama2,bai2023qwen,jiang2023mistral,radford2019language} requires substantial computational resources and data. A major challenge with trained LLMs is their potential to generate inaccurate information in response to user queries. This can occur due to flawed training data or the continuously evolving nature of knowledge \citep{zhang2023large,chen2023combating}.

The increasing popularity of LLMs has highlighted the need for methods to correct factual errors or inaccuracies represented by the models \citep{augenstein2023factuality}. Given the high cost of training LLMs from scratch, recent research has proposed methods for modifying pre-trained LLMs without requiring complete retraining \citep{yao2023editing,wang2023knowledge}. This process, known as "Knowledge Editing" (KE), aims to modify the behavior of pre-trained LLMs to update specific facts without adversely affecting other pre-existing knowledge irrelevant to the requested updates \citep{peng2023check}.

KE presents significant challenges, such as identifying and correcting factual errors within multi-billion parameter LLMs without compromising their overall pre-trained performance. One potential approach for updating an LLM involves naive fine-tuning \citep{wei2021finetuned}, where the parameters of a pre-trained LLM are directly optimized to incorporate new knowledge based on additional data \citep{peng2023check}. However, fine-tuning and even some parameter-efficient fine-tuning (PEFT) methods have drawbacks, including intensive computational requirements, overfitting to the new data, and potential loss of valuable existing knowledge \citep{wang2023knowledge}.

In KE, each factual update, such as changing the pre-trained response "France" to "Argentina" for the question "Who won the FIFA World Cup in Qatar?", is considered a single edit. While KE shares similarities with fine-tuning, it differs by focusing on Locality, Fluency, and Portability metrics, as well as edit accuracy \citep{wang2023knowledge}. Those metrics ensure the model is updated with the new knowledge without degrading the general capabilities of the model. More details about each metric will be provided in Section~\ref{sec:method}. 

We propose a variation of Direct Preference Optimization (DPO) \citep{rafailov2024direct} for KE. This method aims to reduce the likelihood of the model retaining unwanted knowledge while simultaneously increasing the likelihood of incorporating new desired knowledge. Unlike most other methods, our proposed approach does not necessitate additional parameters, external memory, pretraining, or hypernetwork training.
Our contributions are summarized as follows:
(1) We propose viewing KE as an LLM alignment problem. (2) We introduce Knowledge Direct Preference Optimization (KDPO), a variation of DPO that is optimized for incremental knowledge modifications. (3) We conduct extensive empirical experiments on few sequential configurations, across four popular KE datasets, and three popular language evaluation benchmarks, involving multiple LLM architectures demonstrating the advantage of our method. (4) We adapt popular datasets for KE to facilitate sequential editing.

\section{Related Work}
\subsection{Large Model Fine Tuning}

Fine-tuning a large model adapts a pre-trained language model for specific tasks, enhancing its performance on particular datasets. This process aligns the general capabilities of the model with specialized application needs, ensuring more accurate and relevant outputs. However, naive fine-tuning can require significant computational resources and may deviate substantially from the original model. \citet{zhang2023adaptive} proposed Adaptive Budget Allocation for Parameter Efficient Fine-Tuning (AdaLoRA), which allocates computational resources based on the importance of weight matrices.

\citet{christiano2017deep} introduced reinforcement learning from human feedback (RLHF), creating a reward model from human preferences and using reinforcement learning to optimize language model responses. Although effective, RLHF is computationally expensive as it requires training multiple models.

DPO \citep{rafailov2024direct} addresses these challenges by eliminating the need for an additional reward model. DPO uses preference data directly, fine-tuning the language model with token probabilities for chosen and rejected answers, simplifying the process and reducing computational demands.

\subsection{Knowledge Editing}
Knowledge editing in LLMs involves updating or modifying information without retraining the model from scratch. It aims to correct inaccuracies, incorporate new facts, or remove outdated information. This process requires precise adjustments to ensure consistency and reliability across various contexts and queries. 


\citet{mitchell2021fast} introduced Model Editor Networks with Gradient Decomposition (MEND) for quick editing of pre-trained LLMs. MEND uses auxiliary models to convert fine-tuning gradients into efficient weight updates through low-rank gradient decomposition.

\citet{meng2022locating} explored factual storage in GPT models, proposing Rank-One Model Editing (ROME) for precise fact editing with a rank-one MLP update. Later, \citet{meng2022mass} developed Mass-Editing Memory in a Transformer (MEMIT), which scales edits to thousands while preserving model performance.

In a recent study, \citep{zhang2024comprehensive} conducted a survey and presented a new approach. They compared their approach to FT-L, which involves fine-tuning a single layer in a feed-forward network (FFN) based on the ROME algorithm \citep{meng2022locating}. They also introduced FT-M to enhance FT-L, aligning with the fine-tuning objective. FT-M trains the same FFN layer as FT-L using cross-entropy loss on the target answer while masking the original text, leading to state-of-the-art performance on various datasets. In contrast to previous studies that focused on updating the LLM weights, \citep{zheng2023can} proposed In-Context Learning for LLM Knowledge Editing (IKE). They demonstrated competitive results without editing any model weights, which could be particularly valuable in scenarios where model access is restricted.

\section{Method}
\label{sec:method}
\begin{figure*}
\vspace{-0.1 in}
\centering
\includegraphics[scale=0.22]{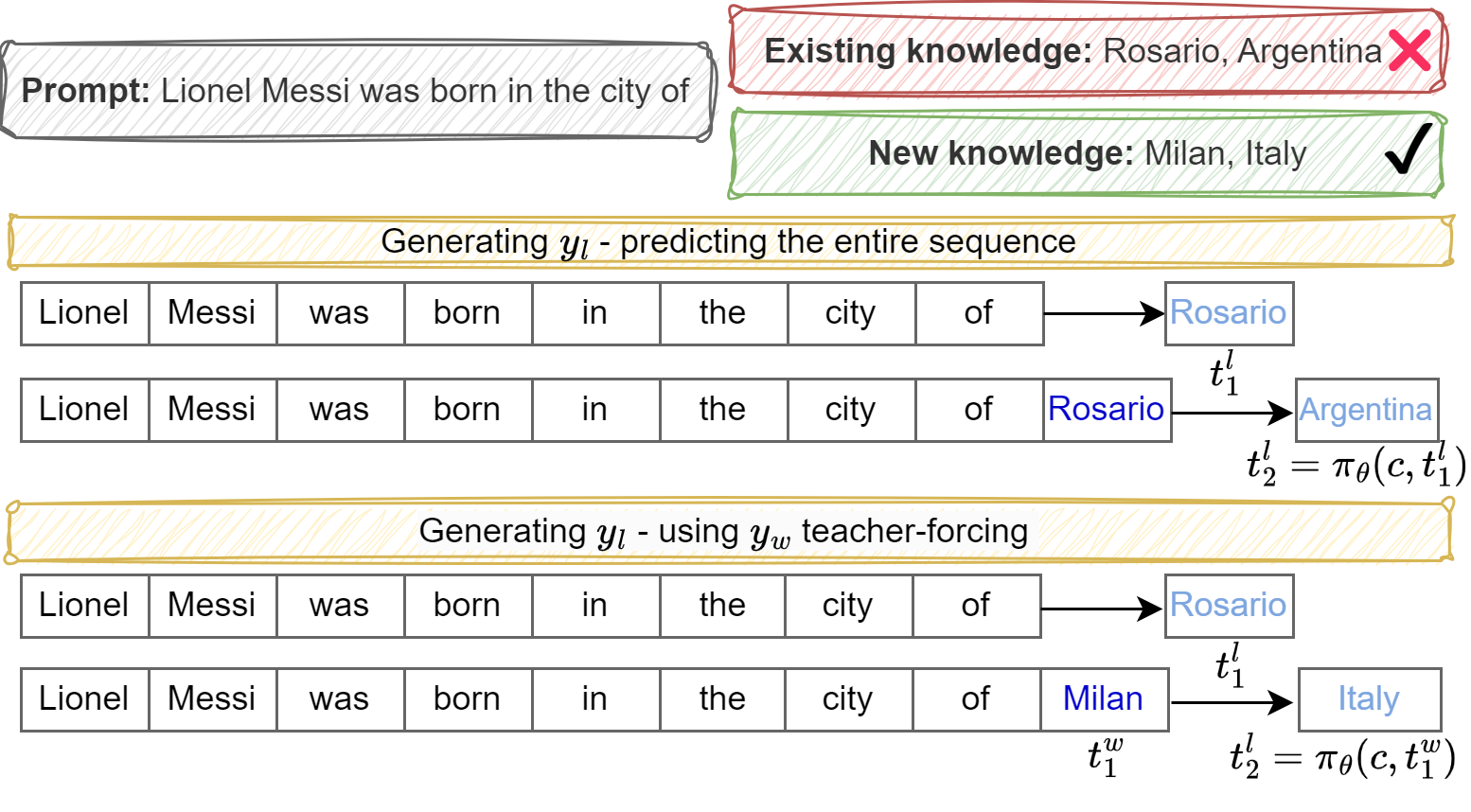}
\caption{Illustration of the two generation methods described in this paper. Words in red are generated based on the previous sequence. In our teacher forcing method, we use the ground truth (in blue) instead of the prediction. This generation will show gains at the optimization step.}
\label{fig:generation}
\end{figure*}
\subsection{Preliminary and Notations}
Let $\mathcal{D} = \{c^i, y^i_w,y^i_l\}_{i=1}^N$ be a dataset composed of $N$ triplets. Each triplet consists of a prompt $c$, a positive completion $y_w$, and a negative completion $y_l$.  
Assume that the vocabulary length of the LLM $\pi_{\theta}$ is $V$. This means that $\pi_{\theta}(c)$ corresponds to a vector of length $V$ and represents the Softmax output of the final layer of the LLM. $y_l$ and $y_w$ can be decomposed into individual token representations: $y_w = (t^w_1, \ldots, t^w_{K_w})$, and $y_l = (t^l_1, \ldots, t^l_{K_l})$, where ${K_w},{K_l}$ are the sequence lengths.

We define, $\pi_{\theta}(y_l | c)$ as the probability of seeing some completion $y_l$ given a prompt $c$. This can be 
 decomposed using the chain rule into a product of probabilities:
\begin{align*}
\pi_{\theta}(y_l | c)= \pi_{\theta}(t^l_k |  c)\prod_{k=2}^{K_l} \pi_{\theta}(t^l_k |  c,\ldots,t_{k-1}^l),
\end{align*}
where $K_l > 2$, in cases $K_l=1$ we get $\pi_{\theta}(t^l_k | c)$.
We will use $c$ as $c_l$ and $c_w$ to clarify the conditioned tokens better.
Thus, $\pi_{\theta}(y_l | c)$ will be represented as $\pi_{\theta}(y_l | c_l)$. For the remainder of the paper, simplified examples will be presented using words as tokens.
\subsection{Knowledge Editing as an Alignment Problem}
The primary objective of LLM alignment is to train models that are safe, effective, ethical, and non-toxic.
LLM alignment is generally performed by finetuning the model using the following objective:
\begin{equation*}
\max_{\pi_{\theta}} \mathbb{E}_{c    \sim \mathcal{D}, y \sim \pi_{\theta}(y|c)} \left[ r_{\phi}(c, y) \right] - \beta \mathbb{D}_{\text{KL}} \left[ \pi_{\theta} \| \pi_{\text{ref}} \right].
\label{eq:alignment}
\end{equation*}
This objective aims to maximize the expected reward $r_{\phi}(c, y)$, with $r_{\phi}$ being a reward model parameterized by $\phi$. Based on user defined criterion, the reward model evaluates the quality of the generated text $y$ given the context $c$.
In our case, we would like $r_{\phi}$ to assign a high reward to the new knowledge we would like to inject into the model.
This alignment objective will aid us in successfully editing the model's knowledge without deviating from the original weights. DPO \citep{rafailov2024direct} showed that it is possible to optimize the same KL-constrained objective without explicitly defining a reward function. Instead, the problem is transformed as a maximum likelihood optimization of the distribution $\pi_{\theta}$ directly, by applying the Bradley-Terry\cite{bradley1952rank} model to the objective $L_{DPO}(\pi_{\theta};\pi_{ref})$, defined as:
\begin{multline*}
-\mathbb{E}_{(x,y_w,y_l) \sim D} 
[ \log \sigma (\beta \log \frac{\pi_{\theta}(y_w | c)}{\pi_{ref}(y_w | c)} - \\ \beta \log \frac{\pi_{\theta}(y_l | c)}{\pi_{ref}(y_l | c)})],
\end{multline*}
where $\sigma$ is the Sigmoid activation. The DPO objective offers two key advantages for KE. First, it employs a coupled preference model, which facilitates the parallel uplifting of $y_w$, the new factual knowledge, and the diminishing of $y_l$, the outdated factual knowledge. Second, the reference model knowledge keeps the model weights from drifting too much, which helps preserve the existing factual knowledge and reasoning capabilities embedded within the model.

\subsection{Knowledge Direct Preference Optimization}
In this section, we present our adapted DPO algorithm for KE, which we term Knowledge Direct Preference Optimization (KDPO). This novel approach differs from vanilla DPO in three key aspects. First, instead of using a pre-prepared preferences dataset, we regularly prompt the model to generate its current knowledge and use the output as the dis-preferred completion, $y_l$. Second, $y_l$ generation is teacher-forced context using $y_w$ as illustrated in Fig.~\ref{fig:generation}. Third, our KDPO optimizes the model using $y_w$ teacher-forced context for the $y_l$ completion, as illustrated in Fig.~\ref{fig:optimization}.

Our optimization is performed using a dataset $\mathcal{D}$, which contains $N$ editing requests. Each request consists of a prompt $c$ and its corresponding new knowledge $y_w$. This can be represented as: $\mathcal{D} = \{c^i, y^i_w\}_{i=1}^N$. We break down the KE process into $n$ cycles, each containing $s$ steps. At the start of a cycle, the model $\pi_{\theta}$ generates a greedy completion $y_l$ to showcase its current knowledge. Subsequently, we optimize $\pi_{\theta}$ over $s$ optimization steps.
Let us further define the context till the $k$-th token: $y_w^{<k}=(t^w_1, \ldots, t^w_{k-1})$ as all tokens in the completion until the $k^{th}$ token.

\subsection{Generation Step}
At the start of each cycle, we initiate the model to articulate its existing knowledge, which is the knowledge we want to erase, denoted as $y_l$. In this setting of KE, we can assume the lengths of $y_w$ and $y_l$ are the same, $K: =K_w=K_l$. The process of generating the $y_l$ can be expressed as follows:
\begin{equation}
y_l = argmax(\pi_{\theta}(c_w)).
\label{eq:single_token_gen}
\end{equation}
It is important to note that the model does not generate the entire answer $y_l$ all at once. Instead, it generates the next token based on the prompt and new knowledge. 
Let us explain why this approach is beneficial. Suppose we have a single sample, $\{c,y_w=(t_1^w,t_2^w,..,t_K^w)\}$, we would like to define $y_l$ that will act as the negative sample in our objective (the definition of which will be provided later). We have two options: (i) predicting the entire sequence and (ii) utilizing teacher-forcing with $y_w$. We describe these options below and illustrate them in Fig.~\ref{fig:generation} using a simple example.

\noindent{\textbf{Predicting the entire sequence}} is done by predicting
$t_1^l=\pi_{\theta}(c)$, then
$t_2^l=\pi_{\theta}(c,t_1^l)$, and continuing iteratively until the last word prediction: $t_K^l=\pi_{\theta}(c,t_1^l,\ldots,t_{K-1}^l)$.
$t_k^l$ are affected by both the prompt $c$ and the generated tokens $y_l^{<k}$, at least in the first cycles, both reflect the original knowledge of the model. 
For example, given the prompt $c=$ "Lionel Messi was born in the city of", and the knowledge we want to embed is $y_w=$"Milan, Italy." The model predicts $t_1^l$="Rosario" based on its current knowledge. Thus, both the prompt and $t_1^l$ (since the model knows Rosario is in Argentina) will cause $t_2^l$ to invoke the model's current knowledge and predict "Argentina".

\noindent{\textbf{New Knowledge Teacher forcing}} is similar to the former only for the first token, which is affected only by the prompt, $t_1^l=\pi_{\theta}(c)$.
The second generated token is affected by the prompt as well as the first word in the new knowledge we desire to embed in the model $t_2^l=\pi_{\theta}(c,t_1^w)$.
The prediction made by the model changes based on its current context which is different at each stage: $c$-original knowledge, and $t_1^w$-part of the new knowledge. Leveraging the previous example, the context for generating $t_2^l$ is now "Lionel Messi was born in the city of Milan". The model's original knowledge will lean the model toward predicting "Argentina", but the "Milan" will push the model to generate "Italy".

Using $t_{k-1}^w$ instead of $t_{k-1}^l$ for generating $t_{k}^l$ will promote a more subtle editing, which is better for ensuring that other parts of the model are unchanged. This will be shown in the following section, which describes our optimization phase.

\subsection{Optimization Step}
Next, let us define the objective for our KDPO, denoted as $\mathcal{L}_{KDPO}(\pi_{\theta}; \pi_{ref})$ using the loss:
\begin{multline*}
     \mathcal{L}_\text{KDPO}(\pi_\theta; \pi_\text{ref})=-\mathbb{E}_{(x,y_w,y_l) \sim D}\\ [\log \sigma (\beta \log \frac{\pi_{\theta}(y_w | c_w)}{\pi_{ref}(y_w | c_w)} - \beta \log \frac{\pi_{\theta}(y_l | c_w)}{\pi_{ref}(y_l | c_w)})].
\label{eq:KDPO}
\end{multline*}
Where, $\pi_{\theta}(y_l | c_w)$ can be decomposed into a product of probabilities:
\begin{align*}
\pi_{\theta}(y_l | c_w)=\prod_{k=1}^K \pi_{\theta}(t^l_k |  c,y_w^{<k}).
\end{align*}
$\pi_{\theta}(t^l_k |  c,y_w^{<k})$ represent the probability of $t_k^l$ appearance based on the prompt and parts of the new knowledge $y_w$. 
KDPO optimizes the model to decrease the probability of the $t_k^l$ with the least amount of bond breaking inside the completion $y_l$. Given the previous example, $y_l$="Rosario, Argentina." DPO will minimize: $\pi_{\theta}(\text{"Argentina"}|\text{"Lionel Messi ... Rosario,"})$  this will aid in breaking the connection between "Lionel Messi" and "Argentina." but it may also detach the relation between "Rosario" and "Argentina."

\noindent{Next,} let us observe how the gradient of $\mathcal{L}_{\text{KDPO}}$ behaves:
\begin{multline*}
\nabla_{\theta} \mathcal{L}_{\text{KDPO}} (\pi_{\theta}; \pi_{\text{ref}}) \propto \\ - [ \nabla_{\theta} \log \pi_{\theta}(y_{\text{w}} | c_w) - \nabla_{\theta} \log \pi_{\theta}(y_{\text{l}} | c_w)],
\end{multline*}
thus
\begin{multline*}
    \nabla_\theta \mathcal{L}_\text{KDPO}(\pi_\theta; \pi_\text{ref}) \propto \\ - \sum_k \nabla_\theta [ \log \pi_\theta(t_k^w|c,y_w^{<k}) - \log \pi_\theta(t_k^l|c,y_w^{<k}) ]. 
\end{multline*}

\begin{figure*}
\vspace{-0.1 in}
\centering
\includegraphics[scale=0.22]{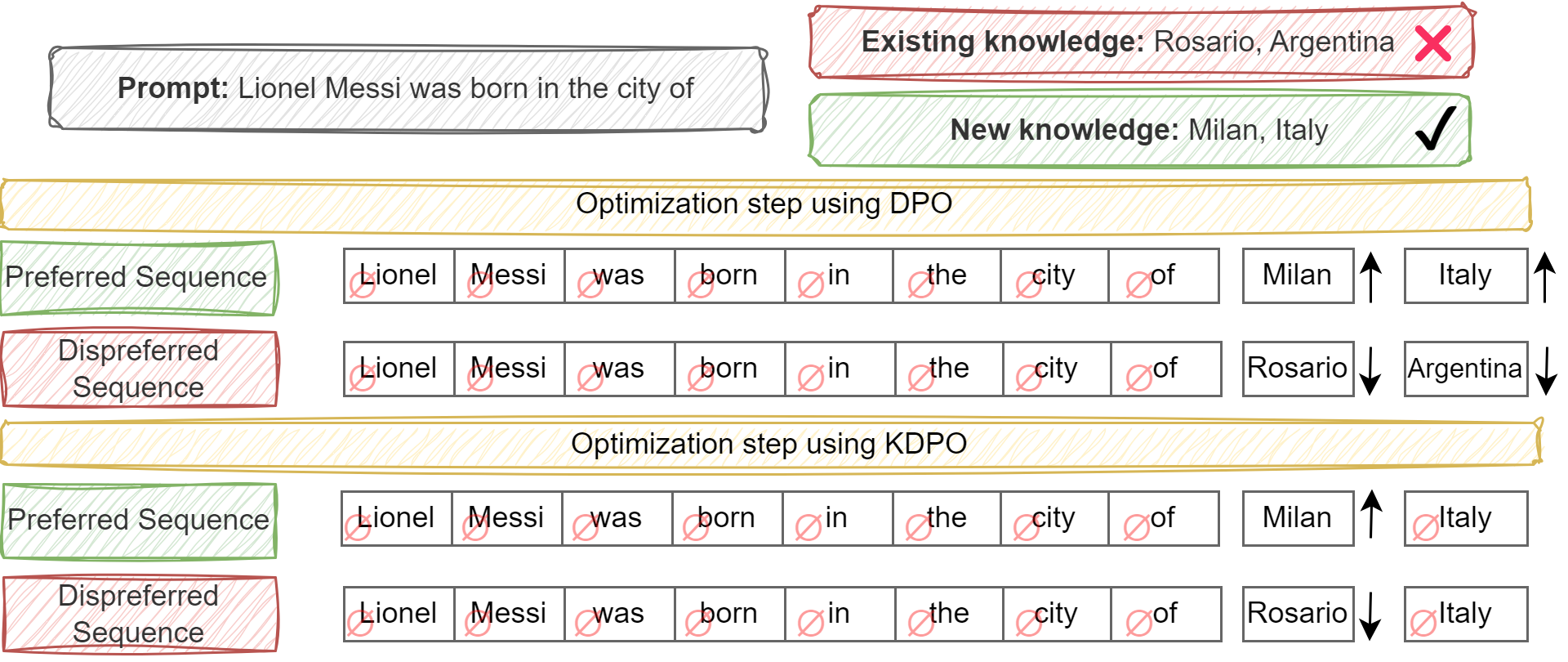}
\caption{Illustration of an optimization cycle. We note tokens that do not affect the loss by the "prohibited" sign. 
The arrows indicate which tokens DPO and KDPO objectives will increase/decrease their log-prob. Note that in KDPO, the "Italy" token is not optimized because it is the same in both sequences, which cancels out the two loss terms of the objective.}
\label{fig:optimization}
\end{figure*}

One can note the importance of generating using $c_w$ and not $c_l$. Using $c_w$, we achieve a more balanced distribution of tokens between the two completions from an early stage; this is beneficial since if $ t_k^l == t_k^w$, the $k$-th term in the sum cancels, which leads to fewer gradient terms to average, thus, fewer changes in the weight space. Fig. 
~\ref{fig:optimization} shows an example of this type of term-cancellation.

\noindent{\textbf{Comparison with FT}}: KDPO enables more reliable KE, by allowing the term-cancellation phenomenon mentioned above. Additionally, because KDPO makes relative adjustments based on a reference model $\pi_{ref}$, it leads to model edits that are more local. Furthermore, KDPO allows for controllable editing capabilities through the use of the $\beta$ parameter. Empirical evidence of this can be seen in Tab.~\ref{tab:100_seq_multi_models}, where the average locality of KDPO is significantly higher than that of FT-M and FT-L (which are fine-tuning variations) for the three models examined.
\begin{table*}[h]
\centering
\begin{adjustbox}{max width=1\textwidth}
\begin{tabular}{@{}l c S S S S S | S S S S S | S  S S S S S@{}}
\toprule
DataSet & \multicolumn{1}{c}{Metric} & \multicolumn{5}{c}{LLaMA3-8B} & \multicolumn{5}{c}{Qwen1.5-7B} & \multicolumn{6}{c}{LLaMA2-7B} \\
\cmidrule(lr){3-7} \cmidrule(lr){8-12} \cmidrule(lr){13-18} 
&{}&  {AdaLoRA}& {ROME} & {FT-L} & {FT-M} & {Ours} & {AdaLoRA}& {ROME} & {FT-L} & {FT-M} & {Ours} & {AdaLoRA} & {ROME} & {MEMIT} & {FT-L} & {FT-M} & {Ours} \\
\midrule
\multirow{4}{*}{ZsRE}
& Edit Succ. & {21.5} & {9.5} & {79.2} & \underline{87.6} & \textbf{88.4} & {29.7}& {51.2} & {49.1} & \underline{70.7} & \textbf{87.1} & {44.7} & {23.1} & {48.3} & {28.9} & \underline{90.8} & \textbf{91.4} \\
& Portability & {11.1} & {3.4}& {26.4}  & \underline{29.9} & \textbf{49.3} & {19.9}& {31.2} & {24.9} & \textbf{46.6} & \underline{42.5} & {28.2} & {7.5} & {24.9} & {9.1} & \underline{48.9} & \textbf{50.4} \\
& Locality & {3.1} & {0.7}& \underline{14.5}  & {4.5} & \textbf{40.4} & {6.3} & {19.5}& {15.5} & \underline{22.2} & \textbf{34.3} & {11.2} & {10.5} & {7.4} & {2.3} & \underline{27.5} & \textbf{45.6} \\
& Fluency & {3.4} & {3.8}& \underline{4.0}  & \underline{4.0} & \textbf{5.0} & {3.4} & \underline{3.2}  & {3.9} & {2.9} & \textbf{5.5} & \underline{4.8} & {4.5} & {4.3} & {2.5} & {3.0} & \textbf{5.5} \\
\midrule
\multirow{3}{*}{WikiBio}
& Edit Succ. & {66.9} & {3.4} & {74.3} & \underline{85.4} & \textbf{89.3} & {76.1} & {64.2}& {55.6} & \underline{87.8} & \textbf{93.5} & {84.4} & {19.4} & {20.3} & {27.1} & \textbf{94.7} & \underline{91.6} \\
& Locality & \underline{16.0} & {6.6} & {9.9} & {10.4}& \textbf{35.8}  & {14.6} & {21.3}& {22.7} & \underline{28.9} & \textbf{36.2} & {20.0} & {9.1} & {8.6} & {7.9} & \underline{36.0} & \textbf{44.4} \\
& Fluency & \textbf{6.3} & {6.0}& {6.1} & \textbf{6.3} & \textbf{6.3} & \textbf{6.3} & {6.0}& {6.0} & \underline{6.2} & \textbf{6.3} & \textbf{6.3} & {6.1} & {6.0} & {5.8} & {6.1} & \textbf{6.3} \\
\midrule
\multirow{4}{*}{Wiki\textsubscript{counterfact}} 
& Edit Succ. & {27.9} & {8.1}& {79.1}  & \textbf{87.4} & \underline{86.7} & {3.2}& {41.2} & {29.8} & \underline{79.6} & \textbf{90.5} & {24.5} & {19.7} & {8.2} & {19.8} & \textbf{92.5} & \textbf{92.5} \\
& Portability & {8.6} & {7.3}& {24.6}  & \underline{28.2} & \textbf{31.2}& {12.1}& {19.8} & {15.8} & \underline{29.2} & \textbf{31.6} & {17.2} & {3.3} & {8.3} & {6.3} & \textbf{48.4} & \underline{47.7} \\
& Locality & {7.6} & \underline{7.9} & {4.1}& {2.7}  & \textbf{42.5}& {10.0}& \underline{52.3} & {48.4} & {48.3} & \textbf{53.9} & {15.6} & {1.9} & {2.4} & {9.1} & \underline{24.5} & \textbf{52.9} \\
& Fluency & {4.0} & \underline{5.0} & {3.7}& {3.8}  & \textbf{5.3} & {3.4} & \underline{4.6}& {4.0} & {2.6} & \textbf{5.5} & \underline{5.3} & {4.6} & {2.7} & {3.9} & {3.2} & \textbf{5.6} \\
\midrule
\multirow{4}{*}{Wiki\textsubscript{recent}}
& Edit Succ. & {12.1} & {4.8}& {78.2}  & \underline{90.4} & \textbf{96.3}&    {48.2}& {71.8} & {62.0} & \underline{81.7} & \textbf{93.3} & {64.2} & {14.9} & {66.9} & {25.8} & \underline{94.4} & \textbf{95.7} \\
& Portability & {4.7} & {5.5} & {31.2}& \underline{36.3}  & \textbf{42.1} & {23.1} & {34.0}& {33.4} & \textbf{39.1} & \underline{37.4} & {39.4} & {5.5} & {30.2} & {9.7} & \underline{54.4} & \textbf{59.0} \\
& Locality & {10.3} & {1.5} & {21.5} & \underline{24.8} & \textbf{47.9}& {24.9}& {38.3} & {49.4} & \textbf{51.2} & \underline{51.0} & {41.9} & {6.6} & {34.1} & {6.5} & \underline{43.2} & \textbf{60.3} \\
& Fluency & {2.7} & \underline{5.2} & {3.3} & {3.2} & \textbf{5.6} & {4.8}& \underline{4.5} & {3.1} & {3.1} & \textbf{5.4} & \textbf{5.6} & {4.5} & {5.0} & {3.4} & {3.6} & \textbf{5.6} \\
\midrule
Average & Edit Succ. & 32.1 & 6.4 & 77.7 & \underline{87.7} & \textbf{90.2} & 39.3 & 57.1 & 49.1 & \underline{80.0} & \textbf{91.1} & 54.5 & 19.3 & 35.9 & 25.4 & \textbf{93.1} & \underline{92.8} \\
& Portability & 8.1 & 5.4 & 27.4 & \underline{31.5} & \textbf{40.9} & 18.4 & 28.3 & 24.7 & \textbf{38.3} & \underline{37.2} & 28.3 & 5.4 & 21.1 & 8.4 & \underline{50.6} & \textbf{52.4} \\
& Locality & 9.2 & 4.2 & \underline{12.5} & 10.6 & \textbf{41.7} & 14.0 & 32.9 & 34.0 & \underline{37.7} & \textbf{43.9} & 22.2 & 7.0 & 13.1 & 6.5 & \underline{32.8} & \textbf{50.8} \\
& Fluency & 4.1 & \underline{5.0} & 4.3 & 4.3 & \textbf{5.6} & 4.5 & \underline{4.6} & 4.2 & 3.7 & \textbf{5.7} & \underline{5.5} & 4.9 & 4.5 & 3.9 & 4.0 & \textbf{5.8} \\
\bottomrule
\end{tabular}
\end{adjustbox}
\caption{Multiple Knowledge Editing algorithm's performance using three different language models (LLaMA3-8B, Qwen1.5-7B, and LLaMA2-7B) on four different datasets (ZsRE, WikiBio, WikiData\textsubscript{counterfact}, and WikiData\textsubscript{recent}) with 100 sequential edits evaluated across multiple metrics. Best result is noted in \textbf{bold} and second best in an \underline{underline}. Overall, our method exhibits good result across models and datasets.}
\label{tab:100_seq_multi_models}
\end{table*}

\begin{algorithm}[tb]
   \caption{Single Edit Flow}
   \label{alg:algorithm}
\begin{algorithmic}
   \STATE {\bfseries Input:} $c$ - context, $y_w$ - new knowledge 
   \STATE {\bfseries Output:} $\theta$
   \STATE {\bfseries Initialize:}  $\pi_{ref} = \pi_{\theta}.copy()$
   \FOR{cycle in n}
      \STATE $y_l = greedy\_generation(\pi_{\theta}(c_w))$ 
      \IF{$y_l == y_w$}
         \STATE break
      \ENDIF
      \FOR{step in s}
         \STATE Calculate loss $\mathcal{L}_\text{KDPO}(\pi_\theta; \pi_\text{ref})$
         \STATE Update $\theta$ using Adam
   \ENDFOR
\ENDFOR
\end{algorithmic}
\end{algorithm}

\section{Experiments}
\label{sec:experiments}
Recent studies \cite{zhang2024comprehensive} have primarily focused on single edits, which involve evaluating a model's performance after a single knowledge update. In contrast, our focus is on sequential editing tasks, which require performing a series of knowledge updates successively, with evaluation conducted after the entire sequence of edits.

\subsection{Evaluation Metrics}
The purpose of KE is to modify the behavior of the model by changing facts. Evaluating KE in LLMs involves assessing the effectiveness and impact of modifications made to the model's knowledge base. This evaluation ensures that the edited model accurately reflects the desired changes while maintaining its overall performance.

However, because facts are interconnected, altering one fact can have unexpected effects on others. This makes it challenging to evaluate edits. To evaluate the capabilities of our framework in KE, we use the same metrics as in previous works \cite{zhang2024comprehensive}, namely edit success, locality, portability, and fluency. Below, we provide definitions of these metrics.

\noindent{\textbf{Edit Success}}: The model should accurately produce updated knowledge for related questions when making an edit. This is evaluated by checking if the post-edit model correctly answers the target knowledge and similar expressions. This builds on previous research \citep{mitchell2021fast,li2024pmet}, combining reliability (exact questions) and generalization (paraphrased questions).

\textbf{Portability}: \citet{yao2023editing} evaluated the model's ability to reason about the implications of the edited knowledge to related content. It is calculated as the average accuracy of the edited model in complex reasoning scenarios. This includes providing the edited continuation when a subject alias or alternative description is used, testing reversed relations, and handling one-hop prompts related to the modification without an explicit edit.

\textbf{Locality}: This metric assesses unrelated knowledge, both in-distribution (e.g., forgetfulness, relation specificity) and out-of-distribution (performance on other NLP benchmarks), to ensure it remains unchanged by the edit algorithm. A good algorithm only affects the intended edits.

\textbf{Fluency}: Measures the diversity and non-repetitiveness of the model's text after editing, using bi-gram and tri-gram entropies. \citet{meng2022locating} suggested lower fluency should be avoided as it indicates the model generates repetitive responses.

\subsection{Datasets}
We assess the performance of our model using four datasets designed to evaluate KE quality: $\textbf{WikiData}_{counterfact}$,\textbf{WikiBio}, \textbf{ZsRE}, and $\textbf{WikiData}_{recent}$. These datasets cover various editing types such as fact manipulation, sentiment modification, and hallucination generation \cite{zhang2024comprehensive}. Details about each dataset are available in Section~\ref{apndx:data_details}. We use the evaluation split provided in EasyEdit \cite{wang2023knowledge}. To prepare these datasets for sequential editing, we have filtered out samples with the same subject. This is to prevent cases where a fact is being edited twice, which would make the first edit non-relevant. For example, if the prompt is "What is the city of birth of X?" and the target is "Y." on the next edit request, "What is the city of birth of X?" with the target "Z." the evaluation performed after $N$ edit requests would be evaluated on answering "Y." even though we have already guided the model's knowledge for another fact.

\subsection{Knowledge Editing Impact on General LLM Tasks}
The efficacy of KE methods is evaluated based on their ability to modify knowledge in the model while preserving its other capabilities, such as reasoning and common sense understanding.  The primary goal is to determine if making targeted factual edits unintentionally hinders the model's capabilities in unrelated areas. To facilitate this analysis, we curate a set of benchmarks HellaSwag \citep{zellers2019hellaswag}, Winogrande \citep{sakaguchi2021winogrande}, and MMLU \citep{hendrycks2020measuring}. Then, we test different KE methods on those benchmarks to investigate any performance degradation. 

\section{Results}
This section presents our results across multiple dataset, LLMs, and KE methods. We start by showing the results of our multiple edits study. Then, we demonstrate our method's ability to handle general, non KE, LLM benchmarks. Lastly, we show an extensive ablation study. Additionally, prompt examples and empirical edit results are presented in Tab.~\ref{tab:example_prompts}. The prompts were selected from the ZsRE dataset to give some intuition on prompt types and edit challenges. We also show comparative responses from multiple KE methods shown in this paper. Some baselines did not converge in all cases and thus were not added to the results. Those issues and other baseline implementation details are discussed in Section~~\ref{subsec:baseimp} of the supplementary material. 
Further results appear in the supplementary material Section~\ref{sec:further_results}.
Our method uses the same hyperparameters for all models (size ranging from 0.5B-8B) and datasets, while we optimized the hyperparameters of baseline methods to achieve competitive results.

\begin{table*}[h!]
\begin{adjustbox}{width=\linewidth,center}
\begin{tabular}{@{}l S S S S S @{}}
\multicolumn{1}{c}{~} & \multicolumn{5}{c}{\textbf{Post-Edit Output}} \\
 \cmidrule(lr){2-6}
{} & {Ours} & {DPO} & {ROME} & {FT-M} & {MEMIT} \\
\midrule
\textbf{Prompt:} Which league did Southern \\ California Fusion join with? \\ \textbf{Pre-Edit Output:} {L W Division} \\ 
\textbf{Edit Target:} {USL First Division} & {USL First Division \cmark} & {USL First Division \cmark} & {~ \xmark} & {USL First Division \cmark}  & {USL  Division \xmark}\\
\midrule
\textbf{Prompt:} What sports team \\ was Petteri Nummelin \\ a member of? \\ \textbf{Pre-Edit Output:} {us Blue Jackbers} \\ 
\textbf{Edit Target:} {Columbus Blue Bombers} & {Columbus Blue Bombers \cmark} & {Olympicus Blue Jackbers \xmark} & {ian \xmark} & {Columbus Blue Bombers \cmark}  & {A  Blue bers \xmark}\\
\midrule
\textbf{Prompt:} The mother of Anthony \\ Delon is whom? \\ \textbf{Pre-Edit Output:} {ida Delon} \\ 
\textbf{Edit Target:} {Alma Delon} & {Alma Delon \cmark} & {Anthonyida Delon \xmark} & {one \xmark} & {Alma Delon \cmark} & {Al Del Jr \xmark} \\
\midrule
\textbf{Prompt:} What network first \\ aired The Smothers \\ Brothers Comedy Hour?\\ 
\textbf{Pre-Edit Output:} {BC} \\
\textbf{Edit Target:} {NBC} & {NBC \cmark} & {NationalBC \xmark} & {BC \xmark} & {NBC \cmark} & {~ \xmark}\\
\midrule
\textbf{Prompt:} What species is ZIC3 \\ specific to? \\  \textbf{Pre-Edit Output:} {~} \\ 
\textbf{Edit Target:} {male} & {male \cmark} & {male} & {male \cmark} & {male \cmark} & {male \cmark}\\
\midrule
\textbf{Prompt:} What war or battle \\ involved Alec Rose? \\  
\textbf{Pre-Edit Output:} { Civil War}  \\
\textbf{Edit Target:} {Spanish Civil War} & {Spanish Civil War \cmark} & {| Civil War \xmark} & {~ \xmark} & {Spanish Civil War \cmark} & {~ \xmark}\\
\midrule
\textbf{Prompt:} What is an ecological \\ status of Bali myna? \\ \textbf{Pre-Edit Output:} {2na} \\ 
\textbf{Edit Target:} {myna} & {myna \cmark} & {crit2na \xmark} & {1udes \xmark} & {1na \xmark} & {myna \cmark}\\
\midrule
\textbf{Prompt:} The father of Juan \\ María Bordaberry is whom? \\ \textbf{Pre-Edit Output:} {inoela Bordaberry} \\
\textbf{Edit Target:} {Gabrielle Bordaberry} & {Gabrielle Bordaberry \cmark} & {Gabrielle Bordaberry \cmark} & {1onú úú \xmark} & {Gabrielle Bordaberry \cmark} & { elle Bordyerry \xmark}\\
\midrule
\end{tabular}
\end{adjustbox}
\caption{Example Pre-Edit and Post-Edit outputs for various prompts taken from the ZsRE dataset. All methods were trained using LLaMA2-7b model using 100 sequential edits. In some cases, the LLM's output was empty either in the pre or post edits. In such cases we leave the answer empty and in the post edit we mark it with an \xmark. Correct responses are marked with a \cmark.}
\label{tab:example_prompts}
\end{table*}
\subsection{Multiple Edits results}
\begin{figure}
\centering
\includegraphics[scale=0.32]{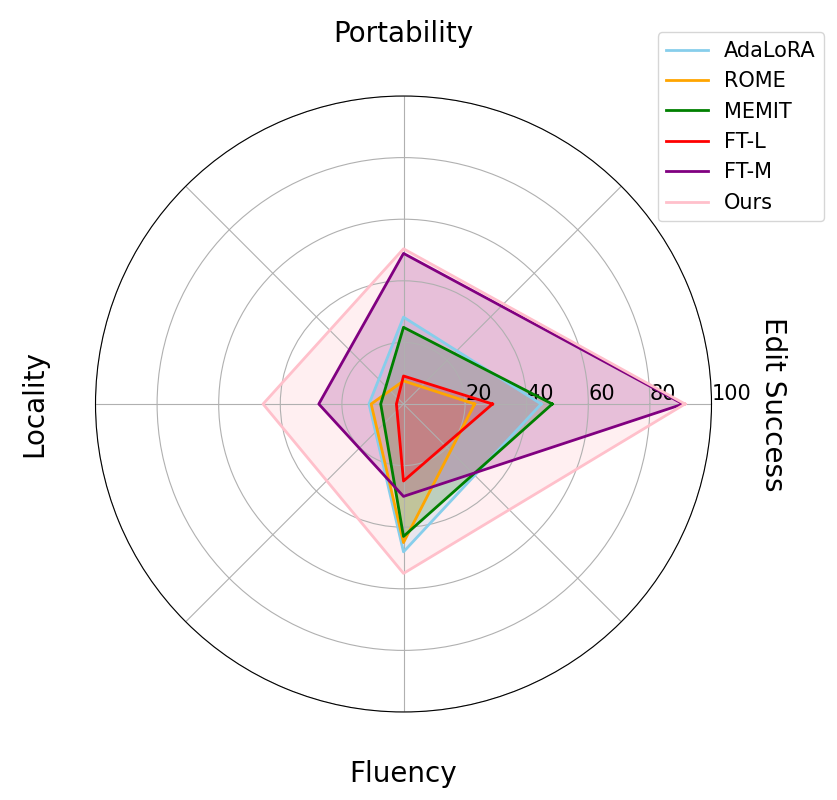}
\vskip -0.15 in
\caption{Comparative result for the four metrics in ZsRE datasets using algorithms discussed in this paper. Fluency results were scaled with a factor of 10 for better visibility.}
\vskip -0.1 in
\label{fig:radar_zsre_llama}
\end{figure}

Fig.~\ref{fig:radar_zsre_llama} compares all KE algorithms on ZsRE dataset using the metrics as described in Section~\ref{sec:experiments}. We show the results for LLaMA2-7B model after 100 sequential edits. Our method outperforms the baselines on all metrics with a notable gap in Locality. This is especially encouraging when using LLMs since we aim to retain the pre-trained knowledge during post-editing. Detailed results for all four datasets for three LLMs (LLaMA3-8B, Qwen1.5-7B, and LLaMA2-7B) are available in Tab.~\ref{tab:100_seq_multi_models} for 100 sequential edits. We show that our method maintains state-of-the-art or comparable results on all datasets in all metrics. In many cases we notice a large gap in the locality, which in some cases surpasses double the performance of other methods. This indicates that our proposed editing method is precise and does not change non relevant parts of the pre trained LLM. To further deepen our understanding of the proposed approach, we tested it on four different LLMs (GPT-j-6B, Qwen1.5-7B, LLaMA2-7B, and LLaMA3-8B) on all datasets using 500 sequential edits in Tab.~\ref{table:500_seq_multi_models}. We notice that performance gap for 500 edit got bigger, our method maintains its performance and achieves state-of-the-art results on all metrics using the recent LLaMA3-8B model.

\begin{table*}[h]
\centering
\begin{adjustbox}{max width=1\textwidth} 
\begin{tabular}{@{}l l |S S S | S  S S | S S S | S S S @{}}
\toprule
DataSet & Metric & \multicolumn{3}{c}{GPT-j-6B} & \multicolumn{3}{c}{Qwen1.5-7B} & \multicolumn{3}{c}{LLaMA2-7b} & \multicolumn{3}{c}{LLaMA3-8b} \\
\cmidrule(lr){3-5} \cmidrule(lr){6-8} \cmidrule(lr){9-11} \cmidrule(lr){12-14}
 & & {AdaLoRA} & {FT-M} & {Ours} & {AdaLoRA} & {FT-M} & {Ours} & {AdaLoRA} & {FT-M} & {Ours} & {AdaLoRA} & {FT-M} & {Ours} \\
\midrule
\multirow{4}{*}{ZsRE} & Edit Succ. &{37.4} & \underline{57.1} &\textbf{69.7} &{43.3} & \underline{66.6} &\textbf{85.4} &{40.6} &\textbf{92.3} &\underline{92.1}& \underline{82.3} &{79.5} &\textbf{87.8} \\
&Portability &\underline{22.9} &{20.8} &\textbf{36.8}&{26.7} &\textbf{46.5} & \underline{41.1}  &{29.8} &\underline{43.8} &\textbf{49.4} & \underline{35.1} &{29.2} &\textbf{44.9}\\
 &Locality & \underline{4.2} & {3.2} & \textbf{22.7} & {5.2} & \underline{18.8} &\textbf{25.0} &{11.8} &\underline{16.4} &\textbf{38.9} & \underline{24.2}& {3.5} &\textbf{31.4}\\
 &Fluency & \textbf{4.4} & \underline{4.2} &{3.8}& \underline{3.1} &{2.6} &\textbf{5.2} & \underline{4.1} & {3.1} &\textbf{5.4} & \underline{4.9} & {4.1} & \textbf{5.4} \\
\midrule
\multirow{3}{*}{WikiBio}& Edit Succ. & \underline{83.1} &  {6.5} & \textbf{88.0} & {76.1} & \underline{87.8} & \textbf{93.5} & {84.4} & \textbf{94.7} & \underline{91.6} & {82.3} & \underline{85.1} & \textbf{87.9}\\
 &Locality&\textbf{23.4} &  {2.1} & \underline{22.8} & {14.6} & \underline{28.9} & \textbf{36.2} & {20.0} & \underline{36.0} & \textbf{44.4} & \underline{25.4} & {13.1} & \textbf{36.2}\\
 &Fluency& \textbf{6.4} &  {5.7} & \underline{6.3} & \textbf{6.3} & {6.2} & \textbf{6.3}  & \textbf{6.3} & {6.1} & \textbf{6.3} & \textbf{6.4} & {6.2} & \textbf{6.4}\\
\midrule
\multirow{4}{*}{Wiki\textsubscript{counterfact}}& Edit Succ.
 & {22.9} & \textbf{66.1} & \underline{65.3} & {24.1} & \underline{74.7} & \textbf{84.5} & {32.8} & \textbf{93.5} & \underline{92.8} & \underline{81.9} &{78.6} & \textbf{84.5}\\
 &Portability& {10.2} & \underline{27.9} & \textbf{29.6} & {11.8} & \textbf{29.4} & \underline{29.1} & {19.5} & \underline{44.4} &\textbf{46.3} & \underline{42.3} & {26.5} & \textbf{60.4}\\
 &Locality& \underline{9.3} & {6.6} & \textbf{20.1} & {8.4} & \underline{43.4} &\textbf{43.6} & {13.7} & \underline{18.1} & \textbf{47.8} & \underline{21.8} & {4.7} & \textbf{32.8} \\
  &Fluency& {3.7} & \textbf{3.9} & \underline{3.8} & \underline{4.3} & {2.9} & \textbf{4.8} & \underline{4.4} & {3.8} & \textbf{5.3} & \underline{5.2} & {4.7} & \textbf{5.4}\\
\midrule
\multirow{4}{*}{Wiki\textsubscript{recent}}& Edit Succ.
 & \underline{58.1} & {57.8} & \textbf{72.2} & {31.4} & \underline{75.7} & \textbf{87.3} & {45.6} & \textbf{93.5} & \underline{92.3} & {81.2} & \underline{83.9} & \textbf{89.5}\\
 &Portability& {3.2} & \underline{26.6} & \textbf{39.9} & {16.7} & \textbf{35.1} & \underline{34.9} & {29.0} & \underline{48.2} & \textbf{52.8} & \underline{31.2} & {29.6} & \textbf{35.9}\\
 &Locality& \underline{31.7} & {17.5} & \textbf{35.4} & {23.4} & \textbf{46.3} & \underline{44.4} & {34.4} & \underline{38.7} & \textbf{50.4} & \underline{28.2} & {24.2} & \textbf{40.3}\\
  &Fluency& \underline{4.1} & \textbf{4.6} & {3.8} & \underline{2.9} & {2.5} &\textbf{5.3} & \underline{3.5} &{3.3} &\textbf{5.6} & \underline{4.2} & {4.0} & \textbf{5.6}\\
\midrule
\multirow{4}{*}{Average} & Edit Succ. & \underline{50.4} & {46.9} & \textbf{73.8} & {43.7} & \underline{76.2} & \textbf{87.7} & {50.9} & \textbf{93.5} & \underline{92.2} & \underline{81.9} & {81.8} & \textbf{87.4} \\
&Portability & {12.1} & \underline{25.1} & \textbf{35.4}& {18.4} & \textbf{37.0} & \underline{35.0} & {26.1} & \underline{45.5} & \textbf{49.5} & \underline{36.2} & {28.4} & \textbf{47.1}\\
 &Locality & \underline{17.2} & {7.4} & \textbf{25.3} & {12.9} & \underline{34.4} & \textbf{37.3} & {20.0} & \underline{27.3} & \textbf{45.4} & \underline{24.9} & {11.4} & \textbf{35.2}\\
 &Fluency & \textbf{4.7} & \underline{4.6} & {4.4}& \underline{4.2} & {3.6} & \textbf{5.4} & \underline{4.6} & {4.1} & \textbf{5.7} & \underline{5.2} & {4.8} & \textbf{5.7} \\
\bottomrule
\end{tabular}
\end{adjustbox}
\caption{Evaluationg of the performance of multiple Knowledge Editing algorithms using four different language models (GPT-j-6B, Qwen1.5-7B, LLaMA2-7B, and LLaMA3-8B) on four different datasets (ZsRE, WikiBio, WikiData\textsubscript{counterfact}, and WikiData\textsubscript{recent}) with 500 sequential edits evaluated across multiple metrics. The best result is noted in \textbf{bold} and second best in an \underline{underline}. Overall, our method exhibits good results across models and datasets.}
\label{table:500_seq_multi_models}
\end{table*}

\subsection{Knowledge Editing in Small Language Model}
Different sizes of language models vary in the way they train, predict, and react to KE. We compare our method to other method as well as to standard DPO. Our method shows very promising results for smaller models like Qwen1.5-0.5B. Fig.~\ref{fig:radar_avg_qwen} shows the average results on three leading datasets . Results are presented in the supplementary material Section~\ref{sec:further_results}, and shows once again that KDPO keeps opens a gap against baseline method on 500 sequential edits. 

\subsection{Knowledge Editing Impact on General LLM Tasks}
We first test the performance of the LLaMA2-7B model on those three benchmarks. Then, we conduct two main experiments to test the performance of LLMs after applying different KE schemes compared to the original LLM. The first experiment tests different KE methods after 100 edits from the ZsRE dataset. Tab.~\ref{tab:genenral_llm_bench} (Left) indicates that our KDPO method performs at a similar level as the pre-trained LLaMA2-7B model, which further strengthens our claim that KDPO possesses strong locality capabilities. On the other hand, the ROME method seems to degrade the performance quite substantially in this case. The second experiment results are in Tab.~\ref{tab:genenral_llm_bench} (Right). This experiment utilizes the WikiData\textsubscript{counterfact} with 500 edits. After making 500 edits to WikiData\textsubscript{counterfact}, we have demonstrated that our method is able to maintain its original performance. Additionally, we have shown that our method outperforms all other tested KE methods across all datasets, sometimes by a significant margin.

\begin{table*}[h]
\begin{minipage}[t]{0.5\textwidth}
    \begin{adjustbox}{width=0.95\textwidth,center}
    \begin{tabular}{l|c|c|c|c}
    \hline
    & HellaSwag & Winogrande & MMLU & Average \\
    \hline
    LLaMA2-7B & {75.99} & {69.06} & 41.24 & 62.76 \\
    \hline
    Ours & \textbf{76.38} & \underline{68.68} & \underline{41.32} & \underline{62.79} \\
    DPO & 75.93 & \textbf{69.29} & \textbf{41.55} & \textbf{62.92} \\
    FT-M & 72.79 & 68.50 & 37.86 & 59.05 \\
    ROME & 27.66 & 48.53 & 24.32 & 33.50 \\
    MEMIT & 72.13 & 66.61 & 26.05 & 54.93 \\
    \hline
    \end{tabular}
    \end{adjustbox}
\end{minipage}
\begin{minipage}[t]{0.5\linewidth}
    \begin{adjustbox}{width=0.95\textwidth,center}
    \begin{tabular}{l|c|c|c|c}
    \hline
    & HellaSwag & Winogrande & MMLU & Average \\
    \hline
    LLaMA2-7B & {75.99} & {69.06} & 41.24 & 62.76 \\
    \hline
    Ours & \textbf{76.28} & \textbf{70.48} & \textbf{38.83} & \textbf{61.86} \\
    DPO & 73.74 & 66.61 & 35.68 & 58.68 \\
    FT-M & 61.55 & 66.37 & 31.93 & 53.95 \\
    ROME & 22.34 & 43.77 & 21.78 & 29.96 \\
    MEMIT & 25.79 & 48.77 & 26.89 & 33.82 \\
    \hline
    \end{tabular}
    \end{adjustbox}
\end{minipage}
\caption{Comparison of performance on general LLM benchmarks (HellaSwag, Winogrande, MMLU). LLaMA2-7B is the base model, with its pre-trained results in the first row for reference. The best results are in bold, and the second-best are underlined. Left: shows KE using the ZSRE dataset for 100 sequential edits. Right: shows KE using the WikiData\textsubscript{counterfact} dataset for 500 sequential edits. Notably, KDPO and DPO are competitive for 100 sequential edits, but KDPO outperforms for 500 sequential edits by a large margin.}
    \label{tab:genenral_llm_bench}
\end{table*}

\subsection{Ablation study}
The goal of this section is to thoroughly examine our proposed method, KDPO versus the Vanilla DPO. We examine both on 100 and 500 sequential edits on various datasets and models. Tab.~\ref{tab:combined_ablation} shows the results of multiple different 6-8B models on three datasets when performing 100 and 500 sequential edits. In the case of 100 edits, the results mostly appear similar, and there seems to be no dominant advantage for our suggested KDPO. However, in the case of 500 sequential edits, KDPO clearly demonstrates its superiority, particularly in the locality metric and the success in edits. We delve deeper into the differences between KDPO and DPO in Tab.~\ref{tab:genenral_llm_bench}. This table illustrate how each method influences the performance of the LLM on general LLM tasks. Our method shows a significantly lower negative impact on the original abilities of the LLM compared to DPO, which should underline the importance of our research findings.

\section{Conclusions}
We have introduced a variant of DPO that is effective for KE. Our extensive testing shows that our proposed KDPO methodology is a promising one for LLM KE. We showed our method works well for various recent LLMs on multiple well known KE datasets. In our ablation, we demonstrated the advatage of KDPO over the vanilla DPO, suggesting the value of our novel idea for KE tasks. Finally, we verified our method maintains the pre trained LLM performance on multiple benchmarks. Overall, we have demonstrated that KDPO is a high-performance and highly precise method for KE tasks. This can significantly help prevent expensive retraining of LLMs due to factual errors.

\section{Limitations}
The primary limitation of our method, which is inherited from DPO, is the need to keep a copy of the model, leading to an increase in the memory footprint. However, it is important to note that recent works, such as those by \citet{meng2024simpo} and \citet{azar2024general}, are actively addressing this challenge, offering potential solutions to reduce this hurdle.

\section*{Acknowledgments}
This work was supported by a grant from the Tel Aviv University Center for AI and Data Science (TAD).

\bibliography{main}
\clearpage
\appendix
\section{Knowledge editing datasets details}
\label{apndx:data_details}
Here we provide the details about KE datasets used in this paper.
\begin{itemize}
\item $\textbf{WikiData}_{recent}$: This dataset focuses on triplets inserted into WikiData after July 2022, enabling creation of insertion edit requests for models trained before this date. These facts are simulating scenarios where an outdated model needs to be updated with new world knowledge.
\item \textbf{ZsRE}: The data involves a context-free question-answering task. In this task, the model is expected to provide the correct object as the answer when given a question based on the subject and relation. We use the extended version, which includes a portability test and uses new locality sets.

\item \textbf{WikiBio}: This dataset aims to correct hallucinations in GPT language models by editing inaccurate sentences from GPT-3 generated Wikipedia-style biographies and replacing them with corresponding sentences from true Wikipedia entries.

\item $\textbf{WikiData}_{counterfact}$: This dataset contains triplets of data about both popular entities (top-viewed Wikipedia pages) and random entities from Wikidata. The random sample is used as the training set, while the popular entities make up the test set. This is because tail entities are often not captured by models and are unsuitable for testing modification edits.
\end{itemize}

\section{Implementation Details}
We used the same hyperparameters for all models across all datasets. We optimized using the Adam optimizer with a learning rate of $1e-4$, number of cycles, $n=$ 10, and the number of steps in each cycle is $s=$ 8. We only optimized layer 21 as done in FT-M. However, empirically examined, the average number of cycles is 4 and the average number of total steps is 29. All experiments were done using PyTorch \citep{NEURIPS2019_9015} on Nvidia A100 GPU.
\subsection{Baseline implementation}
\label{subsec:baseimp}
For the baselines, we used the hyperparameters in the EasyEdit repository. However not all methods contain hyperparameter for all the examined models; In those cases we conduct a grid search on several hyperparameters.
For example, FT-M usually edits layer 31, but Qwen1.5-0.5B has only 24 layers, thus we sweep layers 13-21, layer 15 was the best and that result was presented in the paper.
Further, FT-M mostly use normalization factor of 5e-5, using this factor results in a very poor "edit success" in some models, after searching for better factors, we've used a factor of 5e-4 (only for the models that got poor results using 5e-5).
For some methods we were not able to reach decent results. For example, PMET on LLaMA2-7B did not converge to appropriate numbers, and on the other LLMs we constantly had GPU memory issues. For MEMIT, we were able to reach results on LLaMA2-7B but not on other model due to GPU memory issues. SERAC and MEND we were not able to get reasonable results in training on Qwen1.5 models and LLaMA3-8B.

\label{sec:further_results}

\section{Further Results}
\label{sec:further_results}
\subsection{Small Language Model Results}
\begin{figure}
\centering
\includegraphics[scale=0.32]{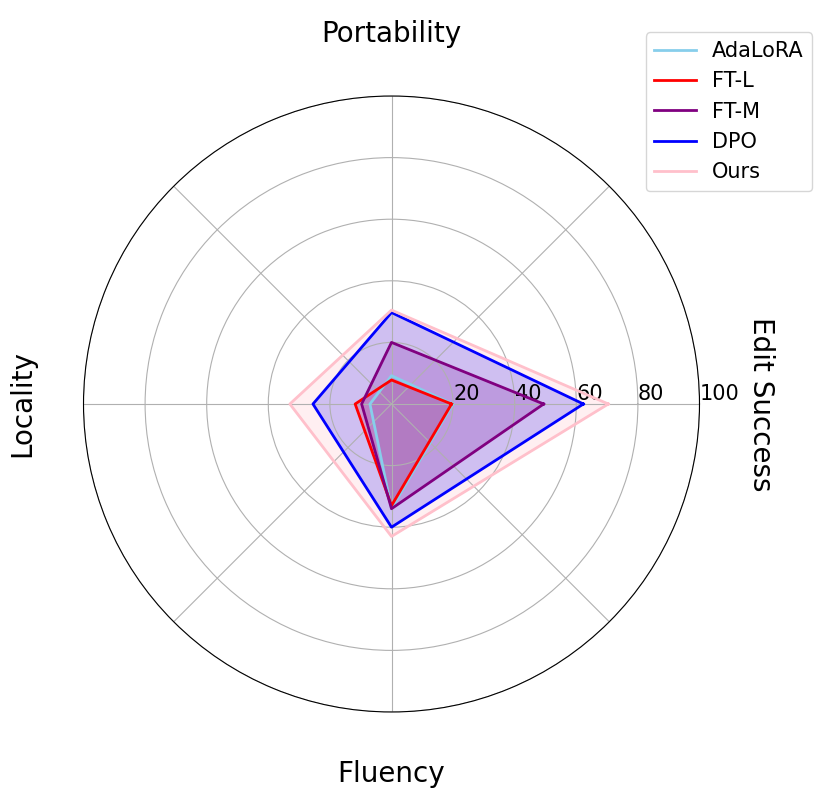}
\caption{Comparative result for the four metrics averaged over all datasets in Tab.~\ref{table:small_lm} using algorithms discussed in this section on Qwen1.5-0.5B with 500 sequential edits. Fluency results were scaled with a factor of 10 for better visibility.}
\label{fig:radar_avg_qwen}
\end{figure}

The results for KE on Qwen1.5-0.5B are present in Tab.~\ref{table:small_lm}. We also show a radar plot of the 500 sequential edits in Fig.~\ref{fig:radar_avg_qwen}. The results shows that the effectiveness gap of our method increases as the number of edits increases, this also can be deduced by other experiments in this paper.
\begin{table*}[h]
\centering
\begin{adjustbox}{max width=1\textwidth} 
\begin{tabular}{@{}l l |S S S S S | S S S S S @{}}
\toprule
DataSet & Metric & \multicolumn{5}{c}{100 sequential edits} & \multicolumn{5}{c}{500 sequential edits} \\
\cmidrule(lr){3-8} \cmidrule(lr){8-12}
 & & {AdaLoRA} & {FT-L} & {FT-M}& {DPO} & {Ours} & {AdaLoRA}& {FT-L}  & {FT-M} & {DPO}  & {Ours} \\
\midrule 
\multirow{4}{*}{ZsRE}
& Edit Succ. & 21.5 & 71.2 & \underline{77.8} & 77.4 & \textbf{85.3} & 20.2 & 20.9 & 43.6 & \underline{66.4} & \textbf{74.3} \\
&Portability & 10.2 & 34.3 & 35.9 & \textbf{43.2} & \underline{43.0} & 10.1 & 8.4 & 20.8 & \textbf{38.0} & \underline{37.7} \\
&Locality & 3.4 & 27.1 & 30.0 & \underline{32.1} & \textbf{34.9} & 3.2 & 4.5 & 4.2 & \underline{24.2} & \textbf{29.9} \\
&Fluency & {3.9} & \textbf{4.4} & \textbf{4.4} & \textbf{4.4} & \textbf{4.4} & 3.6 & 3.3 & 3.0 & \underline{4.4} & \textbf{5.0} \\
\midrule
\multirow{4}{*}{Wiki\textsubscript{counterfact}} 
& Edit Succ. & 13.8 & 54.2 & 63.1 & \underline{69.2} & \textbf{73.3} & 13.2 & 7.6 & 50.3 & \underline{56.1} & \textbf{66.3} \\
&Portability & 5.9 & 21.3 & 23.9 & \underline{26.2} & \textbf{26.8} & 5.5 & 2.3 & 17.8 & \underline{22.6} & \textbf{24.2} \\
&Locality & 4.3 & 29.8 & 35.1 & \textbf{40.1} & \underline{39.1} & 4.0 & 8.4 & 5.8 & \underline{18.8} & \textbf{22.5} \\
&Fluency & 3.1 & \underline{4.2} & \textbf{4.4} & 3.6 & 3.6 & 3.0 & 3.3 & \textbf{3.9} & {3.7} & \textbf{3.9} \\
\midrule
\multirow{4}{*}{Wiki\textsubscript{recent}} 
& Edit Succ. & 27.0 & 64.2 & 76.1 & \underline{79.9} & \textbf{83.5} & 26.3 & 30.1 & 54.4 & \underline{64.3} & \textbf{70.5} \\
&Portability & 13.9 & 31.2 & \textbf{35.6} & \underline{34.4} & \underline{34.4} & 12.1 & 12.6 & 21.4 & \underline{28.5} & \textbf{29.5} \\
&Locality & 14.3 & 51.7 & \textbf{59.1} & 50.3 & \underline{52.4} & 13.8 & 22.6 & 19.0 & \underline{33.4} & \textbf{46.6} \\
&Fluency & 4.3 & {4.6} & \textbf{4.9} & {4.6} & \textbf{4.9} & 3.7 & 3.3 & 3.3 & \underline{3.8} & \textbf{4.0} \\
\midrule
\multirow{4}{*}{Average} 
& Edit Succ. & 20.8 & 63.2 & 72.3 & \underline{75.5} & \textbf{80.7} & 19.9 & 19.5 & 49.4 & \underline{62.3} & \textbf{70.4} \\
&Portability & 10.0 & 28.9 & 31.8 & \underline{34.6} & \textbf{34.7} & 9.2 & 7.8 & 20.0 & \underline{29.7} & \textbf{30.5} \\
&Locality & 7.3 & 36.2 & \underline{41.4} & 40.8 & \textbf{42.1} & 7.0 & 11.8 & 9.7 & \underline{25.5} & \textbf{33.0} \\
&Fluency & 3.8 & \underline{4.4} & \textbf{4.6} & 4.2 & 4.3 & 3.4 & 3.3 & 3.4 & \underline{4.0} & \textbf{4.3} \\
\bottomrule
\end{tabular}
\end{adjustbox}
\caption{Multiple Knowledge Editing algorithm's performance using small language models (Qwen1.5-0.5B) on four different datasets (ZsRE, WikiBio, WikiData\textsubscript{counterfact}, and WikiData\textsubscript{recent}) with 100 and 500 sequential edits evaluated across multiple metrics. The best result is noted in \textbf{bold} and second best in an \underline{underline}. Overall, our method exhibits good results across models and datasets.}
\label{table:small_lm}
\end{table*}
\subsection{Ablation Study Detailed results}
Tab.~\ref{tab:combined_ablation} provides detailed results for our ablation study for 100, and 500 sequential edit on multiple LLM architectures.

\begin{table}[h]
\vspace{0.5in}
\begin{adjustbox}{max width=0.5\textwidth}
\begin{tabular}{l c c c}
\toprule
\textbf{Method} & \textbf{Training} & \textbf{Batch Editing} & \textbf{Weight Update} \\
\midrule
IKE & Yes & No & None \\
AdaLoRA & Yes & Yes & LoRA \\
ROME & Yes & No & Partial \\
FT-L & Yes & Yes & Full \\
FT-M & Yes & Yes & Partial \\
MEMIT & Yes & Yes & Partial \\
DPO & Yes & Yes & Partial \\
Ours & Yes & Yes & Partial \\
\bottomrule
\end{tabular}
\end{adjustbox}
\caption{Comparison of Different Method's Properties}
\label{tab:comparison}
\end{table}

\begin{table*}[h]
\centering
\begin{adjustbox}{max width=1\textwidth} 
\begin{tabular}{@{}l c S S S S S S | S S S S S S S S@{}}
\toprule
DataSet & Metric & \multicolumn{6}{c}{Sequential 100 Edits} & \multicolumn{6}{c}{Sequential 500 Edits} \\
\cmidrule(lr){3-8} \cmidrule(lr){9-16}
 & & \multicolumn{2}{c}{Qwen1.5-7B} & \multicolumn{2}{c}{LLaMA2-7B} & \multicolumn{2}{c}{GPT-j-6B} & \multicolumn{2}{c}{Qwen1.5-7B} & \multicolumn{2}{c}{LLaMA2-7B} & \multicolumn{2}{c}{GPT-j-6B} & \multicolumn{2}{c}{LLaMA3-8B} \\
 & & {DPO} & {Ours} & {DPO} & {Ours} & {DPO} & {Ours} & {DPO} & {Ours} & {DPO} & {Ours} & {DPO} & {Ours} & {DPO} & {Ours} \\
\midrule
\multirow{4}{*}{ZsRE}    
& Edit Succ.    & {92.7}  & {87.1} & {86.4}  & {91.4} & {77.7} & {82.7} & {89.5}  & {85.4} & {81.4}  & {92.3} & {63.6} & {69.7} & {87.8} & {87.8} \\
& Portability   & {45.2}  & {42.5} & {51.1}  & {50.4} & {36.6} & {42.8} & {42.5}  & {41.1} & {44.6}  & {52.8} & {32.7} & {36.8} & {43.7} & {44.9} \\
& Locality      & {35.6}  & {34.3} & {44.8}  & {45.6} & {23.8} & {28.6} & {25.0}  & {25.0} & {41.8}  & {50.5} & {16.0} & {22.7} & {26.4} & {31.4} \\
& Fluency       & {5.5}   & {5.5}  & {5.2}   & {5.5}  & {4.0}  & {3.7}  & {5.2}   & {5.2}  & {4.9}   & {5.6}  & {3.5}  & {3.8}  & {4.7}  & {5.4} \\
\midrule
\multirow{4}{*}{WikiData\textsubscript{counterfact}} 
& Edit Succ.    & {86.8}  & {90.5} & {88.5}  & {92.4} & {75.3} & {75.3} & {77.1}  & {84.5} & {80.6}  & {92.9} & {66.2} & {65.3} & {79.1} & {84.5} \\
& Portability   & {30.8}  & {31.6} & {47.1}  & {47.7} & {33.9} & {34.7} & {27.6}  & {29.1} & {39.9}  & {46.3} & {29.6} & {29.6} & {28.0} & {60.4} \\
& Locality      & {47.6}  & {53.9} & {50.4}  & {52.9} & {22.1} & {24.9} & {20.8}  & {43.6} & {28.0}  & {47.9} & {17.4} & {20.2} & {18.1} & {32.8} \\
& Fluency       & {5.5}   & {5.5}  & {5.3}   & {5.6}  & {4.2}  & {2.9}  & {4.4}   & {4.8}  & {3.9}   & {5.3}  & {3.8}  & {3.8}  & {4.6}  & {5.4} \\
\midrule
\multirow{4}{*}{WikiData\textsubscript{recent}}
& Edit Succ.    & {93.7}  & {93.3} & {97.9}  & {95.7} & {86.8} & {85.5} & {87.1}  & {87.4} & {84.8}  & {92.3} & {71.3} & {72.2} & {89.1} & {89.5} \\
& Portability   & {41.3}  & {37.4} & {60.1}  & {59.0} & {46.3} & {48.2} & {34.8}  & {34.9} & {47.0}  & {52.8} & {37.5} & {39.9} & {37.9} & {35.9} \\
& Locality      & {49.7}  & {51.0} & {61.4}  & {60.3} & {37.7} & {38.8} & {34.9}  & {44.4} & {44.9}  & {50.5} & {32.9} & {35.4} & {41.0} & {40.3} \\
& Fluency       & {5.4}   & {5.4}  & {5.6}   & {5.6}  & {4.6}  & {3.8}  & {5.1}   & {5.3}  & {4.4}   & {5.6}  & {4.5}  & {3.8}  & {4.9}  & {5.6} \\
\bottomrule
\end{tabular}
\end{adjustbox}
\caption{Ablation study: Sequential 100 and 500 edits}
\label{tab:combined_ablation}
\end{table*}
\end{document}